\pdfoutput=1

\documentclass[11pt]{article}

\usepackage[final]{acl}

\usepackage{times}
\usepackage{latexsym}

\usepackage[T1]{fontenc}

\usepackage[utf8]{inputenc}

\usepackage{microtype}

\usepackage{inconsolata}

\usepackage{graphicx}
\usepackage{adjustbox}
\usepackage{amsmath}
\usepackage{amssymb}
\usepackage{amsthm}
\usepackage{array}  
\usepackage{colortbl}
\usepackage{booktabs}
\usepackage{capt-of}
\usepackage[font={small}]{caption}
\usepackage{subcaption}
\usepackage{floatrow}
\usepackage{colortbl}
\usepackage{wrapfig}
\usepackage{multirow}
\usepackage{paralist}
\usepackage{tabularx}
\usepackage{listings}
\usepackage{enumitem}

\usepackage{tikz}
\newcommand*\circled[1]{\tikz[baseline=(char.base)]{
            \node[shape=circle,draw,inner sep=2pt] (char) {#1};}}

\definecolor{mathpurple}{rgb}{0.73, .58, .92}
\definecolor{stringsgreen}{rgb}{.5, .72, 0.42}
\definecolor{entityyellow}{rgb}{1 .83, .15}
\definecolor{relationblue}{rgb}{.42, .64, .82}

\title{Response Wide Shut? Surprising Observations in Basic \\Vision Language Model Capabilities}
\author{
Shivam Chandhok$^{1,2}$ \quad Wan-Cyuan Fan$^{1,2}$ \quad Vered Shwartz$^{1,2,4}$\\ \quad \textbf{Vineeth N Balasubramanian}$^{3,5}$ \quad \textbf{Leonid Sigal}$^{1,2,4}$ \vspace{2mm} \\
$^1$University of British Columbia \quad $^2$Vector Institute for AI \quad
$^3$IIT Hyderabad \quad \\
$^4$CIFAR AI Chair 
\quad
$^5$Microsoft Research India\vspace{1mm}\\
\texttt{\small \{chshivam, wancyuan, vshwartz, lsigal\}@cs.ubc.ca} \quad \texttt{\small vineeth.nb@microsoft.com}
} 

\begin{document}
\maketitle
\begin{abstract}
Vision-language Models (VLMs) have emerged as general-purpose tools for addressing a variety of complex computer vision problems. 
Such models have been shown to be highly capable,
but, at the same time, lacking some basic visual understanding skills.
In this paper, we set out to understand the limitations of SoTA VLMs on fundamental visual tasks by constructing a series of tests that probe which components of design, specifically, may be lacking. 
Importantly, we go significantly beyond the current benchmarks, which simply measure the final performance of VLM response, by also comparing and contrasting it to the performance of probes trained directly on features obtained from the visual encoder, intermediate vision-language projection and LLM-decoder output. In doing so, we uncover shortcomings in VLMs and make a number of important observations about their capabilities, robustness and how they process visual information. We hope our insights will guide progress in further improving VLMs.
\end{abstract}

\section{Introduction}
\label{sec:intro}

Recently, vision-language models (VLMs) have emerged as general-purpose tools that can address many complex language and vision tasks such as comprehending charts and interpreting humor in images and videos \cite{fan2024pretrainingmultimodallanguagemodels,hu2024crackingcodejuxtapositionai,2023videochat,llava,li2023seed, chandhok2024scenegptlanguagemodel3d,talk2bev,wang2024videoagentlongformvideounderstanding}. However, there is also a growing body of evidence that VLMs lack basic capabilities that are considered necessary to solve simple high-level tasks, such as the ability to understand simple negations \cite{Neg} or recognize and count objects \cite{finer,spec2024,VLMClassifier,Paiss2023TeachingCT, chou2025mmr3inconsistencyvisionlanguagemodels}. 
These observations potentially suggest that the mechanisms for solving complex tasks in these models may be different from those in humans, relying more on large-scale matching and memory recall, as opposed to functional and step-by-step reasoning.

\begin{figure}[t]
    \centering
    \includegraphics[width=0.85\linewidth]{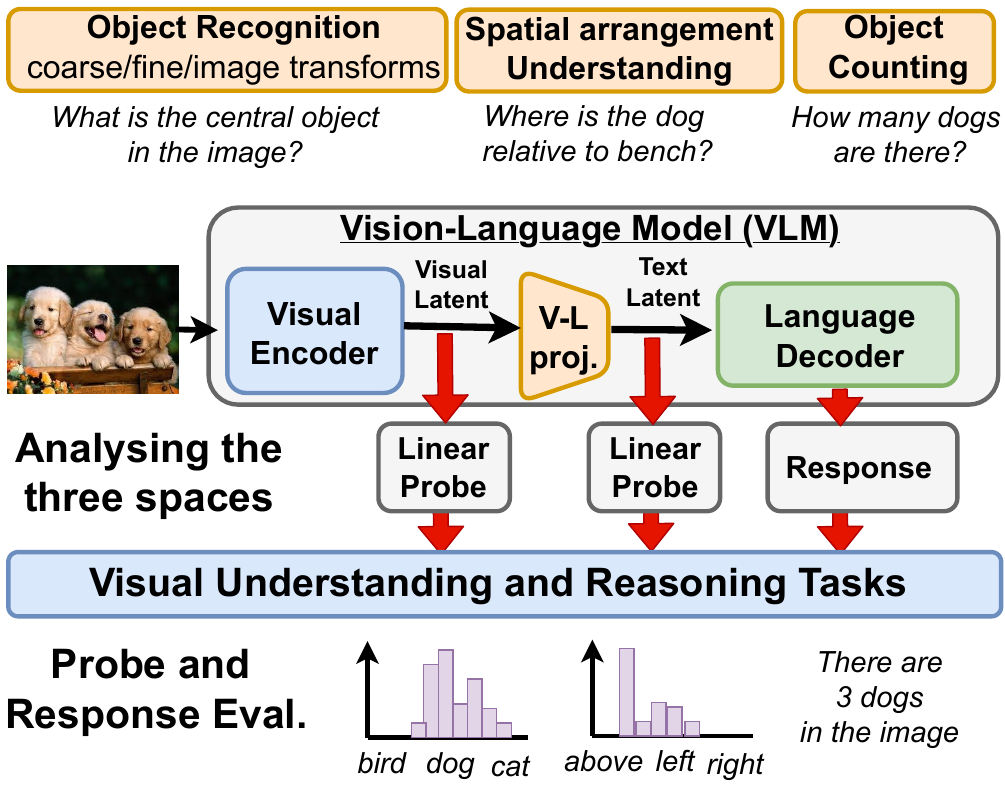}
    \vspace{-0.05in}
    \caption{{\bf Overview of our VLM analysis.} Going beyond existing efforts that analyze VLMs as a whole, we study performance of VLMs in terms of intermediate spaces that represent knowledge as it is processed through the VLM network. Specifically, we consider three spaces in VLMs: \textit{visual},  \textit{VL projection} and \textit{response} space;  to understand what aspects of visual information are captured (not captured) and where.\vspace{-0.2in}} 
    \label{fig:intro_figure}
\end{figure}
\begin{figure*}[t]
    \centering
    \includegraphics[width=0.95\linewidth]{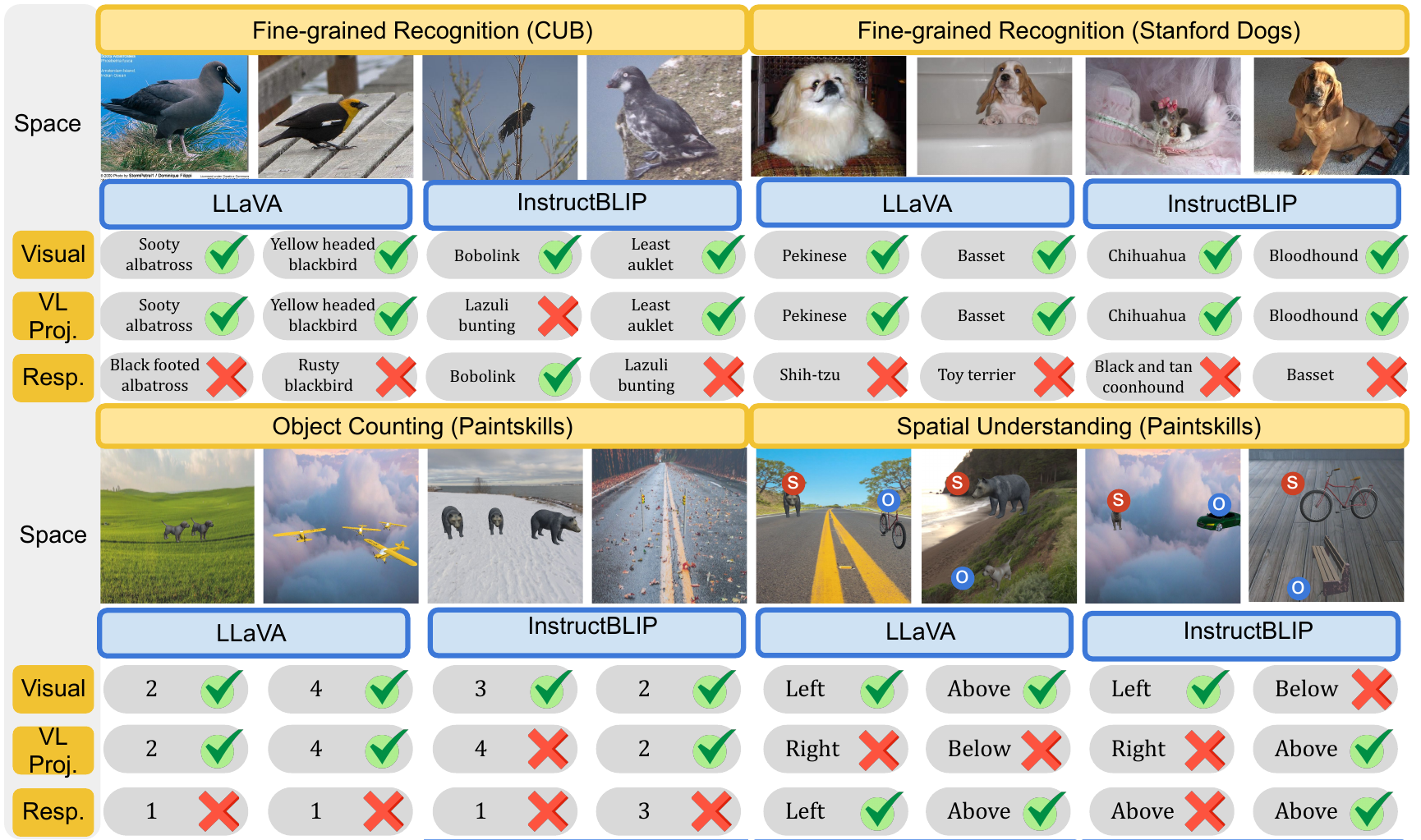} 
    \caption{{\bf Qualitative results supporting the findings of our analysis. } We show prediction (correct vs incorrect) for three spaces i.e \textit{visual,} \textit{VL projection} and \textit{response}. We notice correct predictions in intermediate spaces and incorrect predictions in response space for object recognition and counting task. Furthermore, we notice a reversal in trend for spatial understanding task, where the response space has more correct predictions compared to intermediate spaces.}
    \label{fig:main_figure}
    
\end{figure*}
Inspired by these contradictory observations, we propose a systematic and nuanced analysis of VLMs' performance, focusing on core vision capabilities that we posit are required for high-level visual reasoning tasks: (i) the ability to recognize objects (coarse and fine-grained classification), (ii) delineate instances of a given object type (counting) and (iii) understand their spatial arrangement. 

Previous work that pointed out deficiencies of VLMs on the aforementioned tasks focused only on the final VLM response (the default way to use VLMs) \cite{spec2024,Paiss2023TeachingCT,finer,spatial,VLMClassifier, chou2025mmr3inconsistencyvisionlanguagemodels}. While this allows a holistic measure of performance, it does not provide any insights on which components in VLM design maybe lacking or could be improved.
In this paper, we propose to analyze VLMs' performance \textit{in terms of intermediate spaces that represent information as it is processed through the VLM network} (Figure \ref{fig:intro_figure}). Specifically, VLMs typically consist of a visual encoder, text decoder, and (optional) visual-language (VL)
projection or alignment mechanism. 
Therefore, by analyzing the output of these different modules in their ability to perform visual tasks, we can better understand which capabilities may be missing and where. 
Knowing which of the components is at fault would give important insights into how the performance can be improved and provide a more targetted strategy for enhancing 
VLMs.  

Our observations in this work point to the fact that contrary to current understanding \cite{tong2024eyes}, vision encoders are quite proficient at visual tasks and that the VL projection preserves most (if not all) information. We instead find a significant drop in performance in the VLM response layer---the output of the language decoder, especially in tasks like fine-grained recognition and object counting. 
Figure~\ref{fig:main_figure} shows samples of our qualitative results that examine the predictions after each module: visual, VL proj. and response. We see -- as stated above -- that the final response layer seems to fall short in translating the strong performance of visual and VL projection modules across the recognition and object counting tasks. In  spatial understanding, however, our results reaffirm the results in \cite{tong2024eyes} that the visual and VL proj. space shows weak performance, supporting the need for better visual encoders. 



Beyond these, our studies reveal other interesting observations, w.r.t how background, shape and visual prompting information is processed through spaces within a VLM. Overall, through our insights, our work seeks to identify the key gaps in module performance in VLMs vis-a-vis well-known vision tasks 
and encourage future work to target efforts on improving these gaps in the next generation of vision-language models.


\section{Related Work} 
\label{sec:2}

\noindent\textbf{Vision Language Models (VLMs).} Existing efforts on VLMs can be broadly categorized into three groups based on their training objectives and architectural design: (1) \textit{Contrastive multi-encoder models} such as CLIP \cite{clip} and ALBEF \cite{ALBEF} consist of a separate encoder for each modality ({\em i.e.}, image and text) and use contrastive alignment between image and text inputs for training. 
(2) \textit{Encoder-decoder generative models} such as BLIP-2 \cite{li2023blip2} employ an encoder-decoder architecture to project images to a low-dimensional representation and then use the representation as context to the language decoder to generate language descriptions or captions with generative language modeling loss. 
(3) \textit{Instruction fine-tuned models} such as LLaVA-1.5 \cite{llava}, InstructBLIP \cite{instructblip} and LLaVA-NEXT \cite{llavanext} are additionally fine-tuned to follow human instructions and intent while answering visual questions. Usually, these models are fine-tuned with some form or variant of RLHF \cite{lambert2022illustrating}, or direct preference optimization \cite{Rafailov2023DPO}, which allows them to understand human question intent and answer accordingly. 
In this paper, we analyze VLMs from all three model groups.


\vspace{-0.1in}
\paragraph{Analyzing VLM Capabilities.} 
Recent interests in VLMs have motivated research that aims to understand their visual capabilities. \newcite{distribution}  analyzed how VLMs generalize to
distribution shifts, and 
\newcite{visual_datatypes} further showed a strong correlation between the ability to recognize the type of perturbation and the robustness to it. 
Recently, \newcite{spec2024} evaluated visual-linguistic concepts such as object size, position, existence and count. Similarly, other efforts \cite{spatial,Paiss2023TeachingCT,ARO,Thrush2022WinogroundPV} pointed out deficiencies of VLMs on recognition, spatial, counting etc. However, they only evaluate the final response from the VLMs whereas we test the performance of each component in the VLM individually through our novel three sub-space analysis design to pinpoint where the problem is coming from. Related to our work, \newcite{VLMClassifier} explores image classification in VLMs comparing the performance of LLaVA with contrastive VLMs like CLIP. They find that LLM decoder output (of inference and probe) from  LLaVA \cite{llava} performs inferior to the performance of CLIP. 

In contrast, we go a step further and present an in-depth analysis of VLM's in three intermediate feature spaces that represent the flow of information through the networks allowing us to pinpoint where the deficiencies are coming from. Furthermore, our comprehensive analysis covers a range of diverse tasks in addition to classification, such as counting, spatial orientation, model robustness, prompting and background transformations. 
Additionally, a recent effort, EWS \cite{tong2024eyes}, posit that  deficiencies of the visual encoder affect the VLM's overall performance. Although this is true for spatial tasks, we find that for most visual tasks visual encoders are  proficient/capture necessary information, however, this information is lost in the language decoder and does not translate to an accurate final VLM response.


\section{Experimental Setup}
\label{sec:method}

Motivated by the modular design of VLMs, where individual components are often pre-trained separately and later put together for end-to-end fine-tuning \cite{llava}, we propose to investigate VLMs by looking at their performance in terms of intermediate feature spaces that represent information as it is processed by the network (Fig. ~\ref{fig:intro_figure}). 

Specifically, VLMs typically comprise of a \emph{visual encoder}, \emph{vision-language projection} module and \emph{language decoder} (see Figure~\ref{fig:intro_figure}). This gives rise to 3 different sub-spaces within a VLM: (1) \emph{visual latent} (output of visual encoder) space; (2) \emph{vision-language shared latent} (output of vision-language projection) space, and (3) \emph{language response} space (output of language decoder). We conjuncture that these spaces might capture different aspects of visual information which cumulatively help a VLM understand visual content. 
To this end, we probe these different spaces and analyze their visual capabilities to get a nuanced understanding of what aspects of visual information are captured within VLMs and where. Our design can help pinpoint where the visual knowledge may be lacking or lost,  informing future work  on designing better VLM architectures and training strategies. 

\subsection{Choice of VLMs} 
\label{sec:model_choice}

To comprehensively analyze the visual capabilities of a diverse set of models, we choose representative models from each category (described in Section~\ref{sec:2}): CLIP \cite{clip}, ALBEF \cite{ALBEF} as contrastive multi-encoder models; CoCa \cite{coca} and BLIP-2 \cite{li2023blip2} as encoder-decoder generative models; and InstructBLIP \cite{instructblip}, 
LLaVA v.1.5 \cite{llava}, and LLaVa-NEXT \cite{llavanext} as instruction fine-tuned models. Our choice of these models is also motivated by recent work  on VLM analysis \cite{tong2024eyes,distribution,visual_datatypes}. 
Given the popularity of these models, they are often used as a starting point for 
other more sophisticated VLMs. Hence, analyzing their shortcomings can have a broader impact.\footnote
{At the time of writing, LLaVA-NEXT is widely considered among SoTA \cite{llavanext} open-source models, and the LLaVA-v1.5 model is widely used for downstream applications and fine-tuning VLMs for new tasks.}
Finally, our analysis requires access to the intermediate feature representations, limiting us to open-source models; preventing us from evaluating industrial models such as GPT-4V and Gemini.

\subsection{Choice of Visual Tasks and Benchmarks} 
\label{sec:dataset_choice}

Following prior work, we conjecture that in order to understand the visual content of an image and effectively reason about it, a model must have fundamental capabilities such as the ability to recognize objects ({\em coarse and fine-grained categorization}), group similar object instances together ({\em counting}) and understand the {\em spatial relations} between objects \cite{DallEval}. 
Even though these capabilities are not exhaustive, understanding these dimensions is pivotal for visual reasoning ({\em e.g.}, VQA) as well as building higher-level representations and abstractions like scene graphs \cite{xu2017scenegraph}.

We leverage existing benchmarks for these tasks: 
PaintSkills \cite{DallEval} and Pascal VOC \cite{Everingham10} for coarse object recognition, Stanford Dogs \cite{dogs} and CUB \cite{CUB} for fine-grained recognition,  and PaintSkills \cite{DallEval} for counting (1--4) and spatial reasoning  (above, below, left, right).\footnote{PaintSkills is a diagnostic dataset designed to measure fundamental skills in foundational models.}. We further elaborate on exact details about dataset splits and classes in Appendix \ref{dataset_appendix}  

\begin{table*}[t]
\scriptsize
\centering
\setlength{\tabcolsep}{4pt}
\begin{tabular}{l c cc>{\columncolor{gray!20}}c>{\columncolor{gray!20}}c cc>{\columncolor{gray!20}}c>{\columncolor{gray!20}}c cc>{\columncolor{gray!20}}c>{\columncolor{gray!20}}c}
\toprule
\multirow{2}{*}{VLM Method} & & \multicolumn{4}{c}{PaintSkills} & \multicolumn{4}{c}{Pascal VOC} & \multicolumn{4}{c}{\bf Average} \\ 
\cmidrule(lr){3-6} \cmidrule(lr){7-10} \cmidrule(lr){11-14} 
& & Visual & VL Proj & \multicolumn{2}{c}{Response} 
& Visual & VL Proj & \multicolumn{2}{c}{Response} 
& Visual & VL Proj & \multicolumn{2}{c}{Response} \\
& & {\tt probe} & {\tt probe} & {\tt probe} & {\tt text}
 & {\tt probe} & {\tt probe} & {\tt probe}  & {\tt text}
  & {\tt probe} & {\tt probe} & {\tt probe}  & {\tt text}\\ \midrule
CLIP \cite{clip} & & 99.2 & - & - & 96.5 & 98.5 & - & - & 95.6 & 98.9 & - & - & 96.0 \\
CoCa \cite{coca} & & 99.9 & - & - & 94.0 & 97.6 & - & - & 96.6 & 98.8 & - & - & 95.3 \\ 
ALBEF \cite{ALBEF} & & 99.7 & - & - & 75.3 & 97.3 & - & - & 86.0 & 98.5 & - & - & 80.7 \\ \hline
BLIP-2 \cite{li2023blip2} & 7B & 99.2 & 99.2 & 87.0 & 59.0 & 98.0 & 99.0 & 57.3 & 38.0 & 98.6 &  99.1 &72.2 &{48.5}\\
InstructBLIP \cite{instructblip} & 7B & 99.9 & 99.2 & 98.5 & 98.0 & 98.0 & 98.6 & 95.0 & 70.0 & 99.0 & 98.9 &{96.8} &{84.0} \\ \hline
{LLaVA-1.5} \cite{llava} & 7B & 99.2 & 99.2 & 94.4 & 97.0 & 96.8 & 95.8 & 91.1 & 90.2 & 98.0 & 97.5 & 92.8 & 93.6 \\ 
LLaVA-NEXT \cite{llavanext}& 7B & 99.8 & 99.8 & 94.2 & 97.7 & 97.3 & 96.6 & 94.1 & 94.1 & 98.6 & 98.2 &94.2 &95.9 \\                            
\bottomrule
\end{tabular}
\caption{{\bf Course-grained Recognition.} Visual and VL proj. spaces for all VLMs perform well (accuracy above 95\%). However, there is a small dip in performance in the response space across models (atleast 5\% on average).}
\label{tab:coarse_grained}
\end{table*}

\begin{table*}[t]
\vspace{-0.15in}
\scriptsize
\setlength{\tabcolsep}{4pt}
\centering
\begin{tabular}{l c cc>{\columncolor{gray!20}}c>{\columncolor{gray!20}}c cc>{\columncolor{gray!20}}c>{\columncolor{gray!20}}c cc>{\columncolor{gray!20}}c>{\columncolor{gray!20}}c}
\toprule
\multirow{2}{*}{VLM Method} & & \multicolumn{4}{c}{Stanford Dogs} & \multicolumn{4}{c}{CUB } & \multicolumn{4}{c}{\bf Average} \\ 
\cmidrule(lr){3-6} \cmidrule(lr){7-10} \cmidrule(lr){11-14} 
& & Visual & VL Proj & \multicolumn{2}{c}{Response} 
& Visual & VL Proj & \multicolumn{2}{c}{Response} 
& Visual & VL Proj & \multicolumn{2}{c}{Response} \\
& & {\tt probe} & {\tt probe} & {\tt probe} & {\tt text}
 & {\tt probe} & {\tt probe} & {\tt probe}  & {\tt text}
  & {\tt probe} & {\tt probe} & {\tt probe}  & {\tt text}\\ \midrule
CLIP \cite{clip} & & 94.0 & -& - & 84.0 & 95.0 & -& - & 71.0 & 94.5 & - & -& 77.5\\ 
CoCa \cite{coca} & & 94.0 & -& - & 87.0 & 94.0 & -& - & 74.5 & 94.0 & - & -& 80.8\\ 
ALBEF \cite{ALBEF} & & 83.0 & -& - & 39.0 & 82.5 & -& - & 16.7 & 82.8 & - & -& 27.9\\ \hline
BLIP-2 \cite{li2023blip2}  & 7B & 93.6 & 92.0 & 58.1 & 23.5 & 92.0 & 93.2 & 20.5 & 10.0 & 92.8 &92.6 &39.3 &16.8 \\
InstructBLIP \cite{instructblip}  & 7B & 93.6 & 92.6 & 28.5 & 12.0 & 92.2 & 94.0 & 60.7 & 13.0 &92.9 & 93.3 & 44.6 & 12.5 \\ \hline
{LLaVA-1.5 }\cite{llava} & 7B & 91.9 & 88.7 & 25.9 & 19.9 & 92.2 & 87.3 & 34.1 & 34.6 & 92.1 & 88.0 & 30.0 & 27.3 \\
{LLaVA-NEXT }\cite{llavanext} & 7B & 90.4 & 87.2 & 37.5 & 31.0 & 90.0 & 85.0 & 22.5 & 18.0 & 90.2 & 86.1 & 30.0 & 24.5 \\                            
\bottomrule
\end{tabular}
\caption{{\bf Fine-grained Recognition.} 
There is significant drop in performance in the response space for fine- vs. course-grained recognition. In the visual and VL projection feature spaces the drop also exists, but is significantly less, comparatively speaking.}
\label{tab:fine_grained}
\vspace{-0.1in}
\end{table*}

\vspace{-0.05in}
\subsection{Task Design}
\label{sec:task_design}

We formulate the aforementioned tasks as classification over a set of discrete choices ({\em e.g.}, for {\em classification} -- a set of coarse/fine object classes; for {\em counting} -- a set of counts; and {\em spatial orientations} -- a set of spatial arrangements, such as above, below, left-of, right-of). 
We convert these classification tasks to a question form
 to test visual understanding  capabilities of VLM models \cite[as used by previous work:][]{finer,distribution} where the VLM is prompted with a VQA query such as ``{\em What is the central object in the image? Choices - dog, cat, ...?}''.
We further elaborate on task prompts and details of probing in Appendix~\ref{task_prompts}.

%

\vspace{-0.05in}
\subsection{Evaluation Design}
\paragraph{Visual and VL Projection Space.} To evaluate the visual and VL projection spaces, we probe the frozen feature representations by training task-specific linear probes.  Following previous work \cite{clip, dino, probing3d}, we use the standard linear probing strategy of a single layer (MLP) logistic regression model.
Specifically, we train a probe on an average pooled features from output of vision encoder and VL proj. for each image. The probes are trained on training split and evaluated on val/test splits.


\paragraph{Response Space.} We evaluate the response space in two ways: (1) for fair comparison with the visual and VL projection, we similarly train a linear probe on the output token representation from the language decoder (\texttt{Probe}).  
Specifically, the output of the language decoder for a given  image is of the form $T \times F$, where $T$ is the number of tokens and $F$ is the feature dimension of token embeddings,  which we average pool to produce $F$-dimensional feature for each image. 
(2) we directly evaluate the textual response of the VLM to a VQA query for a given task such as ``{\em What is the central object in the image? Choices - dog, cat, ...?}'' (\texttt{Text}). 
Note that this is the default way to evaluate VLMs.\footnote{CLIP, CoCa, and ALBEF don't have a language decoder, hence we only report the \texttt{Probe} for these models.}
(See Appendix~\ref{task_prompts} for of task prompts and evaluation)

\paragraph{Control Task for Probes.}
To verify that representation encodes  properties useful for the task rather than the linear probe learning the task from the training data, we include a  
control task
in which we randomly shuffle the labels \cite{zhang-bowman-2018-language, control_task}. Good performance on the target task combined with bad performance on the control tasks validates our linear probing results can serve as proxy for information present (See Appendix~\ref{probe_design} for the control task results).

\section{Analyzing the Skills of VLMs}
\label{sec:results}

\subsection{Object Recognition} 
\label{sec:objrec}

\noindent\textbf{(Coarse-grained) object recognition.} Table~\ref{tab:coarse_grained} presents the results of VLMs on coarse-grained object recognition. 
The visual and VL projection spaces for all VLMs perform well, 
resulting in accuracy above 95\%. However, there is some dip in performance in the response space across models (atleast 5\% on average). Even though all models show similar high performance in visual and VL proj. space, BLIP family of models perform worse in the response space compared to other models. We attribute this gap to the fact that BLIP-2 wasn't instruction tuned which may prevent it from understanding the instructions in the query even when it otherwise has access to the required information in the visual and projection space. Evidence for this can be seen by comparing BLIP-2 to its instruction-tuned counterpart  -- InstructBLIP which results in a substantial performance increase in response space. However, InstructBLIP still does not reach the performance of LLaVA-1.5 and LLaVA-NEXT on Pascal VOC, which we attribute to stronger and better instruction-tuning for the latter models. The better performance of the linear prob ({\tt probe}) compared to ({\tt text}) space on InstructBLIP also supports this hypothesis; {\em i.e.}, it is able to capture the answer, but not verbalize it in ({\tt text}). 

\begin{figure}[t]
    \vspace{-0.1in}
    \centering
    \begin{subfigure}[b]{0.49\textwidth}
\includegraphics[width=\textwidth]{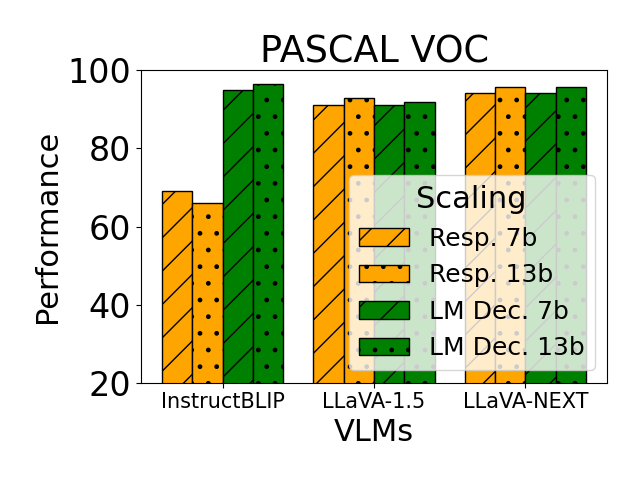} 
        
        \label{fig:plot1}
    \end{subfigure}
    \hfill
    \begin{subfigure}[b]{0.49\textwidth}
\includegraphics[width=\textwidth]{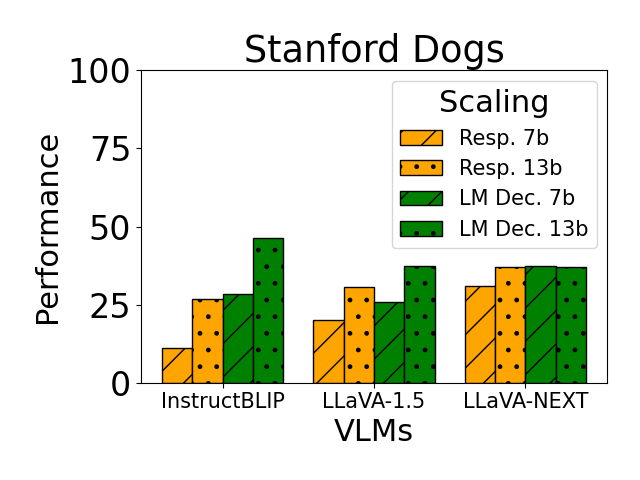} 
        
        \label{fig:plot2}
    \end{subfigure}
    
    \caption{\textbf{Effect of Scaling LLM Decoder on Response space.} Refer to Appendix \ref{scaling_appendix} for elaborate results\vspace{-0.2in}}
    \vspace{-0.1in}
    \label{fig:scaling}
    
\end{figure}

\noindent\textbf{(Fine-grained) object recognition.} Table~\ref{tab:fine_grained} presents the results of VLMs on fine-grained object recognition. 
The decrease in performance in the response space compared to the visual and VL projection spaces is drastic (atleast 45 \% drop), 
across both datasets. Our experiments point to problems in the response space by showing that the visual and VL projection probes perform very well, with most models achieving an accuracy above 90\% across datasets. However, the knowledge does not translate to the response space. These results shed light on the findings in \newcite{finer} that VLMs are overall less capable in fine-grained recognition, and contradict the conclusion of \newcite{tong2024eyes} that the deficiency is due to the visual encoder.  
Further, we note that \newcite{finer} show that text-only language decoders are themselves good at fine-grained object classification. Hence, we conjecture that the drastic dip in performance is not due to lack of knowledge in the language decoder; rather it is due to ineffective joint fine-tuning of the proj. layer and language decoder
which is responsible for aligning the vision and language modalities. 

\begin{table}[t]
    \scriptsize
    \setlength{\tabcolsep}{4pt}
    \centering
    \begin{tabular}{@{}lccc>{\columncolor{gray!20}}c>{\columncolor{gray!20}}c@{}}
    \toprule
    \multirow{2}{*}{VLM Method} & & \multicolumn{3}{c}{PASCAL VOC}  \\ 
    \cmidrule(lr){3-6}
& & Visual & VL Proj & \multicolumn{2}{c}{Response} \\
& & {\tt probe} & {\tt probe} & {\tt probe} &  {\tt text} \\ \midrule 
    LLaVA-NEXT \cite{llavanext}   & 7B &  97.3 &96.6 &94.1 &94.1 \\
    LLaVA-NEXT \cite{llavanext}   & 13B &  97.3 &96.2 &95.7 &95.8 \\
    LLaVA-NEXT \cite{llavanext}   & 34B &  97.4	&96.5	&97.0	&97.0 \\
    \bottomrule
\end{tabular}
\vspace{-0.2in}
\label{table:scaling_pascal}
\end{table}

\begin{table}[t]
    \scriptsize
    \setlength{\tabcolsep}{4pt}
    \centering
    \begin{tabular}{@{}lccc>{\columncolor{gray!20}}c>{\columncolor{gray!20}}c@{}}
    \toprule
    \multirow{2}{*}{VLM Method} & & \multicolumn{3}{c}{Stanford Dogs}  \\ 
    \cmidrule(lr){3-6}
& & Visual & VL Proj & \multicolumn{2}{c}{Response} \\
& & {\tt probe} & {\tt probe} & {\tt probe} &  {\tt text} \\ \midrule 
    LLaVA-NEXT \cite{llavanext}   & 7B &  90.4 &87.2 &37.5 &31.0 \\
    LLaVA-NEXT \cite{llavanext}   & 13B &  90.5 &86.5 &37.0 &36.9 \\
    LLaVA-NEXT \cite{llavanext}   & 34B &  91.0	&86.9	&51.3	&49.4 \\
    \bottomrule
\end{tabular}
\caption{\textbf{Effect of Scaling LLM Decoder.}  Results are shown on three spaces for LLaVA-NEXT on coarse-grained (top) and fine-grained (bottom) recognition task. \vspace{-0.2in}} 
\label{table:scaling_pascal}
\end{table}
\noindent\textbf{Object recognition discussion and implications.} While the encoding of the fine-grained classes could be seen as being preserved in the visual and intermediate VL projection space, it is clear that representations in that space do not align well with those in the language decoder. We investigate this further and find that the LLaVA 665k fine-tuning data has only 0.17\% of samples about fine-grained dog breeds. Given our findings, we posit that the data used for joint fine-tuning of VLMs' projection with language decoder did not contain enough samples (was not representative) of fine-grained classes.
Hence, incorporating fine-grained samples and improving joint fine-tuning of the VL proj. and language decoder would yield improvements in fine-grained  recognition.

\vspace{0.1in}
\noindent\textbf{Impact of model size.} An important question to ask is if the drop in performance in the response space is due to the size of the language decoder. Figure~\ref{fig:scaling}, Table~\ref{table:scaling_pascal} and Appendix\ref{scaling_appendix} show the effect of scaling the language decoder on accuracy. We observe that a bigger LM decoder does generally improve the performance of response space to some extent, including improvements in the VL proj. space. However, the best response accuracy still remains atleast 35\% lower than the corresponding visual and VL proj. space performance for fine-grained object recognition.



\begin{table}[t]
    \scriptsize
    \setlength{\tabcolsep}{4pt}
    \centering
    \begin{tabular}{@{}lccc>{\columncolor{gray!20}}c>{\columncolor{gray!20}}c@{}}
    \toprule
    \multirow{2}{*}{VLM Method} & & \multicolumn{3}{c}{PaintSkills}  \\ 
    \cmidrule(lr){3-6}
& & Visual & VL Proj & \multicolumn{2}{c}{Response} \\
& & {\tt probe} & {\tt probe} & {\tt probe} &  {\tt text} \\ \midrule 
    CLIP \cite{clip} & & 93.5 & - & - & 49.0 \\ 
    CoCa \cite{coca} & & 91.7 & - & - & 77.0 \\ 
    ALBEF \cite{ALBEF} & & 78.1 & - & - & 40.0 \\ \midrule
    BLIP-2 \cite{li2023blip2} & 7B &  96.6 & 95.3 & 26.0 & 25.0 \\
    InstructBLIP \cite{instructblip}   & 7B &  96.6 & 95.6 & 82.0 & 82.0 \\ \midrule
    \multirow{2}{*}{LLaVA-1.5 \cite{llava}} & 7B &  93.0 & 94.0  & 81.0 & 81.0 \\ 
                                        & 13B & 93.1 &  94.0 & 73.0 & 73.8 \\
    LLaVA-NEXT \cite{llavanext}   & 7B &  94.4 & 95.7 & 81.2 & 81.2 \\
    \bottomrule
\end{tabular}
\caption{\textbf{Object Counting on PaintSkills.} See text for detailed analysis and discussion.} 
\label{table:counting}
\end{table}
\begin{table}[t]
    \scriptsize
    \centering
    \setlength{\tabcolsep}{4pt}
\begin{tabular}{@{}lccc>{\columncolor{gray!20}}c>{\columncolor{gray!20}}c@{}}
\toprule
    \multirow{2}{*}{VLM Method} & & \multicolumn{3}{c}{PaintSkills}  \\ 
    \cmidrule(lr){3-6}
& & Visual & VL Proj & \multicolumn{2}{c}{Response} \\
& & {\tt probe} & {\tt probe} & {\tt probe} &  {\tt text} \\ \midrule 
    CLIP \cite{clip}  & & 47.5 & - & - &  31.0 \\ 
    CoCa \cite{coca}  & & 48.0 & - & - &  27.3 \\ 
    ALBEF \cite{ALBEF}  & & 47.0 & - & - &  28.0 \\ \midrule
    BLIP-2 \cite{li2023blip2}  & 7B & 50.4 & 50.0 & 27.7 & 25.0 \\
    InstructBLIP \cite{instructblip}    & 7B & 50.4 & 57.0 & 27.7 & 26.8 \\ \midrule
    \multirow{2}{*}{LLaVA-1.5 \cite{llava}} & 7B &  50.2 &  51.0 &  61.0 &  63.0 \\ 
                                        & 13B & 50.5 &  49.0 &  72.6&  74.0 \\  
    LLaVA-NEXT \cite{llavanext}    & 7B &  49.8 & 49.8 & 37.6 & 37.6 \\
    \bottomrule
\end{tabular} 
\caption{\textbf{Spatial Understanding on PaintSkills.} See text for detailed analysis and discussion.\vspace{-0.2in}} 
\label{table:spatial}
\end{table}

\subsection{Counting and Spatial Understanding}
\label{sec:counting}

Next, we consider object counting and spatial arrangement understanding more difficult than object recognition, which is a prerequisite for both.
Following \newcite{DallEval}, we use the PaintSkills diagnostic evaluation dataset for both tasks, as it allows us to control object placement and arrangement. Specifically,  we use images with 1--4 instances for the counting  and 4 orientations ({\em left}, {\em right}, {\em above}, {\em below}) for spatial arrangement task. The chance 
performance on both tasks is 25\%.

\noindent\textbf{Object counting.} The results reported in Table~\ref{table:counting} illustrate trends similar to those observed for fine-grained recognition, albeit less pronounced. Mainly, the performance of the visual and VL projection probes is high ($>$90\%), with the {\em VL projection} representations marginally trailing {\em visual}, while performance in the response space is considerably poorer (a drop of at least 14\%). 

\vspace{0.1in}
\noindent\textbf{Spatial understanding.} Table~\ref{table:spatial} shows the trend observed thus far reverses on spatial understanding: the performance of {\em visual} latent representation is the worst; it increases a little for the {\em VL projection} space (for InstructBLIP and LLaVA-1.5), and increases further in the response space (for LLaVA-1.5). We hypothesize that the superior performance of the response space compared to visual space is due to the small number of samples pertaining to spatial arrangement in the training dataset for the visual encoder.
LAION-2B used by OpenCLIP contains only 0.2\% of spatial samples \cite{spatial}. In contrast, to \textasciitilde 35\% in LLaVA-665k used by LLaVA-1.5 to fine-tune response space. Additionally, lower performance of LLaVA-NEXT response compared to LLaVA-1.5, could be due to relative drop of relevant data from 35\% in LLaVA-665k to \textasciitilde 26\% in data used by LLaVA-NEXT.  

Based on these results, we make a few important deductions. First, the spatial understanding capabilities of VLMs are clearly inferior, with the visual encoder ({\em i.e.}, CLIP) being responsible for the loss of information. Second, it appears that for some VLMs ({\em e.g.}, LLaVA-1.5) the response space makes up for some loss of performance thanks to the relevance of training data. 
These findings are in line with previous work showing that the performance of VLM on spatial understanding and counting \cite{spatial,Paiss2023TeachingCT} 
highly depend on the number of samples which are representative of these tasks in fine-tuning data. 

\begin{table}[t]
    \scriptsize
    \setlength{\tabcolsep}{3pt}
    \centering
    \begin{tabular}{@{}lccc>{\columncolor{gray!20}}c@{}}
    \toprule
    \multirow{2}{*}{Dataset/Task} & & \multicolumn{3}{c}{Performance}  \\ 
    \cmidrule(lr){3-5}
& & LLaVA-1.5 & (blind) Vicuna LLM &Chance \\
& & {\tt response(text)} & {\tt response(text)} &   {\tt} \\ \midrule 
    PASCAL (Coarse-Cls.)  & &  90.2	&8.7	&6.6  \\
    CUB (Fine-Cls.)   & &  34.6	&7.7	&6.6  \\
    Dogs (Fine-Cls.)   & &  19.9	&11.6	&6.6	 \\
    PaintSkills (Spatial)   & &  63.0	&25.6	&25.0	 \\
    \bottomrule
\end{tabular}
\caption{\textbf{Impact of language priors.}  We compare performance of LLaVA-1.5 (VLM) with (blind) LLM to isolate role of language priors\vspace{-0.2in}}
\label{table:blindcomparison}
\end{table}

\noindent\textbf{Impact of language priors.} Prior work has shown that language priors can influence the responses generated by VLMs \citep{goyal2017makingvvqamatter,langpriors_1,langpriors_2}. For example, VLMs may learn from their language priors that the most likely spatial arrangement of a man and a chair is that the man is on the chair. These models are often able to solve VQA tasks ``blindly'' -- {\em i.e.}, without an image encoder \cite{langpriors_1}, or when the objects of interest are masked \cite{langpriors_2} -- based on language priors. 
To reduce such confounding effects, our setup and datasets are carefully selected to minimize the influence of language priors. In particular, datasets like PaintSkills are synthetically constructed with uniformly distributed object placements, avoiding spurious correlations between language and visual content. 
Finally, we also present results from a ``blind'' language-only LLM baseline,  
further isolating the role of language priors in VLM performance. Specifically, we compare the text-only responses from instruction-tuned blind LLM (Vicuna) with those of LLaVA-1.5, enabling a clearer assessment of how much the VLM's performance depends on true visual reasoning versus language biases (Table~\ref{table:blindcomparison}). The low accuracy of the (blind) Vicuna LLM indicates the contribution of the language prior is relatively insignificant. These findings are consistent with prior work showing that language priors in generative VLMs -- similar to those used in our setup -- are generally not strong enough to have substantial positive impact on performance of visual tasks \citep{langpriors_2}.

\begin{figure}[t]
    \centering
    \begin{subfigure}[b]{0.45\linewidth}
        \centering
        \includegraphics[width=\linewidth]{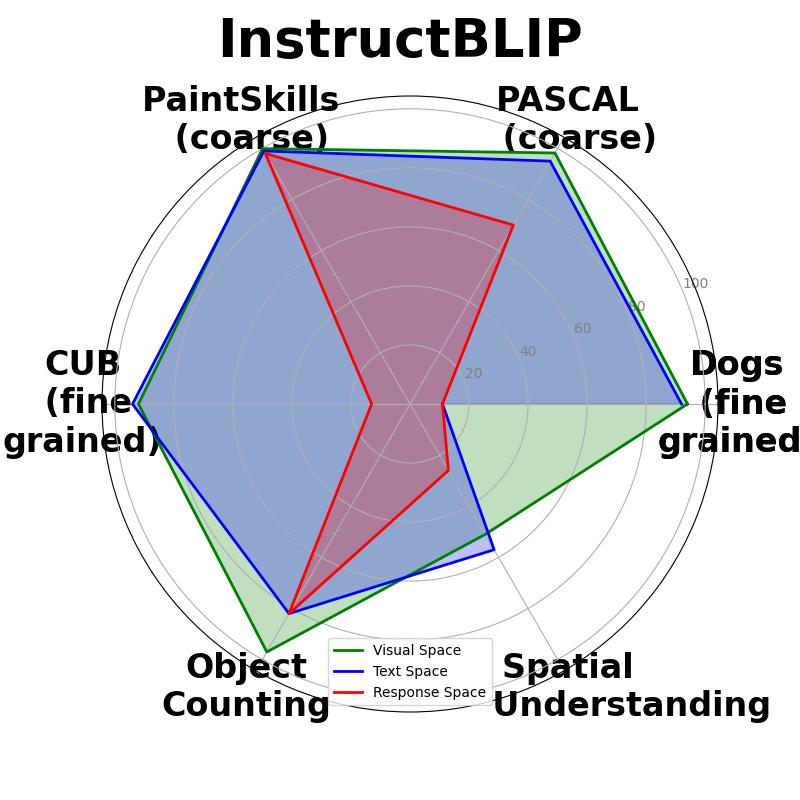} 
        \label{fig:intro_figure_1}
    \end{subfigure}
    \hspace{0.05\linewidth} 
    \begin{subfigure}[b]{0.45\linewidth}
        \centering
        \includegraphics[width=\linewidth]{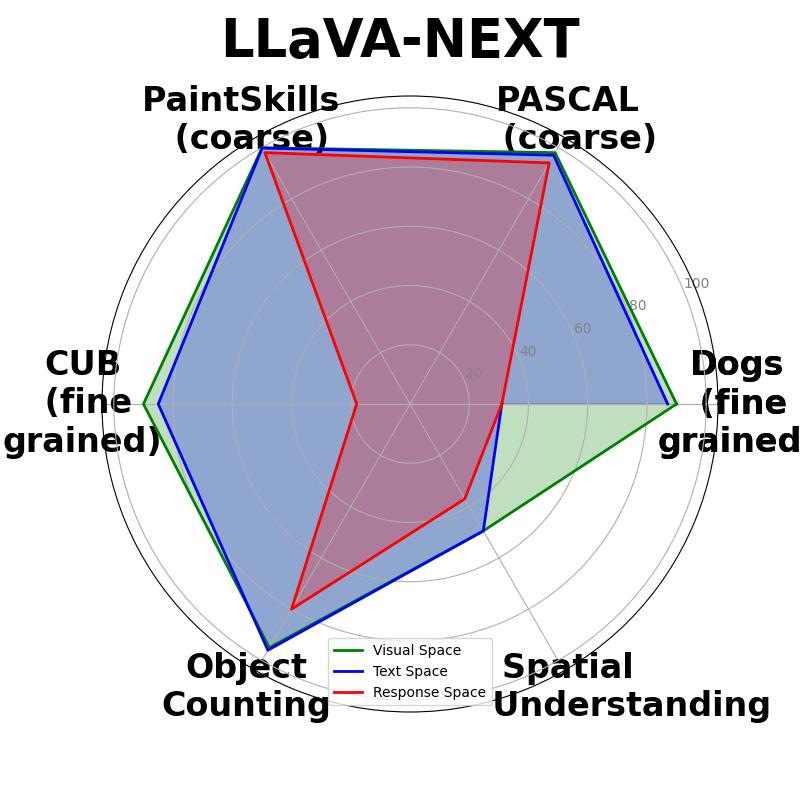} 
        \label{fig:intro_figure_2}
    \end{subfigure}
        \vspace{-0.3in}
    \caption{{\bf Consolidation of performance} for instruction-tuned models from the BLIP family and the LLaVA family. Legend: {\color{green}Visual space}, {\color{blue}Text space}, {\color{red}Response space}. \vspace{-0.2in}}
    
    \label{fig:side_by_side_figures}
\end{figure}

\subsection{Summary of Insights on VLM Skills}
\label{sec:results_summary}

The summary of all experiments in this section is compactly illustrated for two of the models (InstructBLIP and LLaVA-NEXT) in Figure~\ref{fig:side_by_side_figures}. 
\begin{enumerate}[leftmargin=*,topsep=4pt]
\setlength\itemsep{0em}
    \item[\scriptsize \circled{1}] VLMs, overall, are much less capable of recognizing fine-grained categories which is largely attributable to response from language decoder. Knowledge from visual and VL proj. spaces does not translate to the VLM final response due to ineffective joint fine-tuning stage of the proj. and language decoder. 

    \item[\scriptsize \circled{2}]Scaling the language decoder does not solve the aforementioned problem and the drop in performance in
the response space remains significant.
    \item[\scriptsize \circled{3}] The substantial drop in response space performance on counting is attributable to the limitations of the language decoder; similar to observation 1 above.
    \item[\scriptsize \circled{4}]We observe a significant lack of ability of vision encoders to capture arrangement information, which results in a reduced performance overall for spatial understanding tasks. 
\end{enumerate}


\section{Analyzing the Robustness of VLMs}
\label{sec:analysis}
Here we explore another dimension of visual understanding in VLMs: robustness to (1) visual corruptions (\S\ref{sec:analysis:visual}), and (2) background changes (\S\ref{sec:analysis:background}). 

\newcommand{\subred}[1]{\textsubscript{{\color{red}#1}}}
\newcommand{\subgain}[1]{\textsubscript{{\color{blue}#1}}}

\begin{table}[t]
\scriptsize
\setlength{\tabcolsep}{5pt}
\centering
\begin{tabular}{l l ll>{\columncolor{gray!20}}l>{\columncolor{gray!20}}l }
\toprule
& & Visual & VL Proj & \multicolumn{2}{c}{Response} \\
VLM Method & Data & {\tt probe} & {\tt probe} & {\tt probe} &  {\tt text} \\ \midrule 
\multirow{2}{*}{BLIP-2} & Orig    & 
98.0  & 99.0 & 57.6 & 38.0   \\
                                         & Corrupt &  95.2\subred{-2.8}  & 95.4\subred{-3.6} & 57.5\subred{-0.1} & 38.2\subgain{+0.2}   \\ 
                                         \hline
\multirow{2}{*}{InstructBLIP}    & 
Orig    &98.0  &98.6 &95.0 &70.0    \\ 
                                         & Corrupt &95.3\subred{-2.7} &95.9\subred{-2.7} &90.5\subred{-4.5} &69.5\subred{-0.5}  \\ 
                                         \hline
\multirow{2}{*}{LLaVA-1.5}      & Orig
&96.8  &95.8 &91.0 &90.2    \\ 
                                         & Corrupt &90.8\subred{-6.0}  &89.3\subred{-6.5} &87.7\subred{-3.3} &87.4 \subred{-2.8}  \\ 
                                                   \hline
\multirow{2}{*}{LLaVA-NEXT}     & Orig 
& 97.3  &96.6 &94.1 &94.1   \\ 
                                         & Corrupt &89.7\subred{-7.6}  &88.2\subred{-8.4} &87.3\subred{-6.8} &87.3\subred{-6.8}  \\ 
\bottomrule
\end{tabular}
\caption{\textbf{Robustness to Visual Corruptions.} Robustness experiment on the Pascal VOC dataset. 
\vspace{-0.2in}}
\label{tab:robust}
\end{table}

\begin{table*}[t]
\setlength{\tabcolsep}{3pt}
\scriptsize
\centering
\begin{tabular}{l cc>{\columncolor{gray!20}}c>{\columncolor{gray!20}}c cc>{\columncolor{gray!20}}c>{\columncolor{gray!20}}c cc>{\columncolor{gray!20}}c>{\columncolor{gray!20}}c}
\toprule
\multirow{2}{*}{Transformation} &  \multicolumn{4}{c}{LLaVA-NEXT \cite{llavanext}} & \multicolumn{4}{c}{LLaVA-1.5 \cite{llava}} & \multicolumn{4}{c}{InstructBLIP \cite{instructblip}} \\ 
\cmidrule(lr){2-5} \cmidrule(lr){6-9} \cmidrule(lr){10-12} 
& Visual & VL Proj & \multicolumn{2}{c}{Response} 
& Visual & VL Proj & \multicolumn{2}{c}{Response} 
& Visual & VL Proj & \multicolumn{2}{c}{Response} \\
& {\tt probe} & {\tt probe} & {\tt probe} & {\tt text}
 & {\tt probe} & {\tt probe} & {\tt probe}  & {\tt text}
  & {\tt probe} & {\tt probe} & {\tt probe}  & {\tt text}\\ \midrule
(1) Original    & 77.5 & 74.5 & 67.9 & 67.5 & 76.9 & 74.9 & 68.7 & 68.1 & 77.5 &78.5 &62.7 &60.0  \\ \midrule
(2) Black BG    & 
        88.3\subgain{+10.8} & 
        87.0\subgain{+12.5} & 
        73.2\subgain{+5.3} & 
        72.5\subgain{+5.0} & 
        87.6\subgain{+10.7} & 
        87.1\subgain{+12.2} & 
        73.6\subgain{+4.9} & 
        72.5\subgain{+4.4} & 
        90.5\subgain{+13.0} & 
        90.7\subgain{+12.2} & 
        71.4\subgain{+8.7} & 
        61.0\subgain{+1.0} \\ 
        
(3) White BG    & 
        88.2\subgain{+10.7} & 
        86.7\subgain{+12.2} & 
        73.0\subgain{+5.1} & 
        72.3\subgain{+4.8} & 
        88.7\subgain{+11.8} & 
        87.3\subgain{+12.4} & 
        74.2\subgain{+5.5} & 
        73.0\subgain{+4.9} & 
        90.4\subgain{+12.9} & 
        89.2\subgain{+10.7} & 
        70.6\subgain{+7.9} & 
        56.1\subred{-3.9} \\ 
(4) Silhouette + Orig BG  & 
        63.7\subred{-13.8} & 
        59.7\subred{-14.8} & 
        44.4\subred{-23.5} & 
        41.1\subred{-26.4} & 
        60.0\subred{-16.9} & 
        57.8\subred{-17.1} & 
        45.7\subred{-23.0} & 
        42.1\subred{-26.0} & 
        69.3\subred{-8.2} & 
        69.4\subred{-9.1} & 
        39.8\subred{-22.9} & 
        41.7\subred{-18.3}   \\
(5) Silhouette + White BG & 
        47.9\subred{-29.6} & 
        44.7\subred{-29.8} & 
        22.1\subred{-45.8} & 
        20.3\subred{-47.2} & 
        44.3\subred{-32.6} & 
        42.8\subred{-32.1} & 
        23.3\subred{-45.4} & 
        21.4\subred{-46.7} & 
        51.2\subred{-26.3} & 
        46.9\subred{-31.6} & 
        24.5\subred{-38.2} & 
        19.4\subred{-40.6} \\ \midrule
(6) Blur Reverse & 
        87.8\subgain{+10.3} & 
        87.5\subgain{+13.0} & 
        72.7\subgain{+4.8} & 
        74.2\subgain{+6.7} & 
        86.9\subgain{+10.0} & 
        85.3\subgain{+10.4} & 
        75.9\subgain{+7.2} & 
        75.0\subgain{+6.9} & 
        90.9\subgain{+13.4} & 
        92.0\subgain{+13.5} & 
        75.8\subgain{+13.1} & 
        64.2\subgain{+4.2} \\   
(7) Red Circle + Orig BG  & 
        81.8\subgain{+4.3} & 
        78.1\subgain{+3.6} & 
        71.1\subgain{+3.2} & 
        70.7\subgain{+3.2} & 
        80.3\subgain{+3.4} & 
        77.3\subgain{+2.4} & 
        72.9\subgain{+4.2} & 
        72.1\subgain{+4.0} & 
        82.4\subgain{+4.9} & 
        85.0\subgain{+6.5} & 
        69.5\subgain{+6.8} & 
        64.8\subgain{+4.8} \\
(8) Red Circle + White BG & 
        86.5\subgain{+9.0} & 
        85.5\subgain{+11.0} & 
        72.2\subgain{+4.3} & 
        71.5\subgain{+4.0} & 
        86.5\subgain{+9.6} & 
        85.5\subgain{+10.6} & 
        73.0\subgain{+4.3} & 
        72.0\subgain{+3.9} & 
        89.0\subgain{+11.5} & 
        88.2\subgain{+9.7} & 
        71.5\subgain{+8.8} & 
        62.1\subgain{+2.1} \\ \midrule
(10) Edge & 
        62.0\subred{-15.5} & 
        60.1\subred{-14.4} & 
        53.1\subred{-14.8} & 
        52.4\subred{-15.1} & 
        61.3\subred{-15.6} & 
        58.8\subred{-16.1} & 
        54.2\subred{-14.5} & 
        53.2\subred{-14.9} & 
        70.0\subred{-7.5} & 
        68.4\subred{-10.1} & 
        53.5\subred{-9.2} & 
        50.1\subred{-9.9} \\
(11) Patch Shuffle  & 
        73.5\subred{-4.0} & 
        70.1\subred{-4.4} & 
        65.1\subred{-2.8} & 
        65.2\subred{-2.3} & 
        72.5\subred{-4.4} & 
        70.4\subred{-4.5} & 
        67.2\subred{-1.5} & 
        66.5\subred{-1.6} & 
        75.8\subred{-1.7} & 
        75.1\subred{-3.4} & 
        54.0\subred{-8.7} & 
        53.0\subred{-7.0} \\
\bottomrule
\end{tabular}
\caption{\textbf{Background Transformation Experiments.} Owing to limited space, we present results on a subset of models.\vspace{-0.2in}} 
\label{tab:background}
 \vspace{-0.1in}
\end{table*}

\subsection{Visual Corruptions}
\label{sec:analysis:visual}

 VLMs are fairly robust to many common corruptions, 
presumably due to the prevalence of these effects in the large-scale training data \cite{visual_datatypes, distribution}. We thus focus 
on a diverse set of corruptions by which they are more affected. Specifically, we include noise (Gaussian, uniform, shot, and  impulse), blurs (motion, defocus) and weather changes (snow) following \newcite{corruption}. We use methodology in \newcite{benchmark_corruption} and \newcite{corruption}, and add corruptions to Pascal VOC. 

\vspace{0.1in}
\noindent\textbf{Robustness Across Spaces.} 
 Table~\ref{tab:robust} shows results for robustness across spaces. Here we report the average performance across corruptions (Avg. Corrupt), the original performance without corruptions (Orig) and their difference (in red). 
 Detailed per-corruption breakdown is in 
 Appendix~\ref{visual_corr_appendix}. 
 
We observe that despite its deficiencies (see \S\ref{sec:results}), the response space is the most robust among the three spaces we consider. Interestingly, in line with our findings that visual information does not transfer well to the response space, we see that the effects of the corruptions did not translate to the response space as well, making it \emph{appear} as the most robust space.
Furthermore, we observe that the intermediate VL projection latent space is less robust than the visual and response space, especially for LLaVA-1.5 and LLaVA-NEXT---pointing to deficiency in the VL projection. 
Compared to LLaVA family, the VL proj. is more robust for the BLIP family, which use a Q-former and cross-attention for mapping visual information to language modality, making VL proj. more stable than LLaVA which simply uses single layer MLP. 

\subsection{Background Transformations}
\label{sec:analysis:background}

Here, we explore how VLMs process foreground vs. background information. Additionally, inspired by recent work that applies background transformations through visual prompting to enhance VLM performance \cite{fineprompt,red_circle}, we explore how these techniques impact intermediate spaces. While prior work only focus on CLIP-like contrastive models, we explore how VLMs like LLaVA, InstructBLIP process background and visual prompting information through intermediate spaces.  The sample constructed stimuli are illustrated in Figure~\ref{fig:bg}. We use COCO \cite{background} and provide details of how we applied the transformations in Appendix~\ref{background_implementation}. The results of this experiment are summarized in Table~\ref{tab:background}. Overall, we observe that for most visual prompting transformations, the gains are higher in the visual and text space and relatively diminish in the response space. Targeted efforts to reduce information loss in the response space should increase the VLM response further for visual prompting.

\begin{figure}[!t]
\begin{center}
    \vspace{-0.15in}
    \setlength{\tabcolsep}{3pt}
    \begin{tabular}{cccc} 
    {\tiny ~} & & & \\ 
        \includegraphics[width=0.21\textwidth]{rws_images/None.png} 
        & \includegraphics[width=0.21\textwidth]{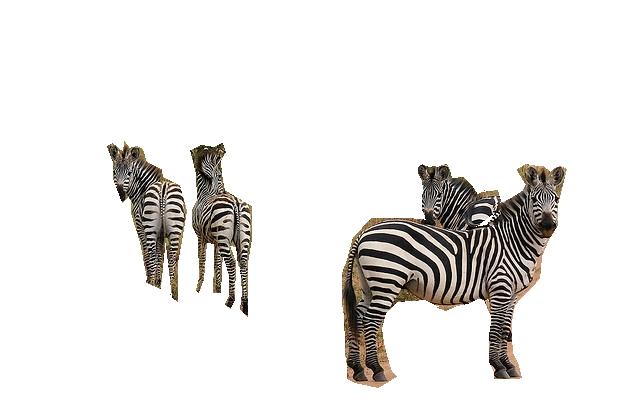} 
        &\includegraphics[width=0.21\textwidth]{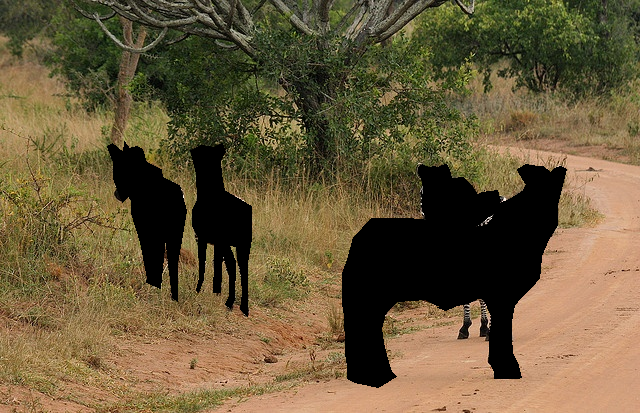} 
        &\includegraphics[width=0.21\textwidth]{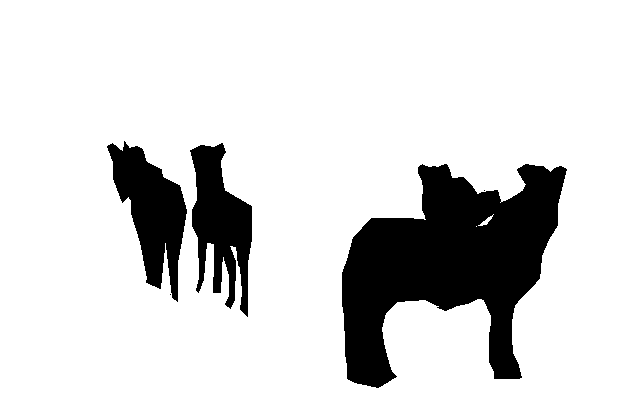}
        \\ 
        {\small \it Original} & {\small \it White BG} & {\small \it Silhouette} & {\small \it Silhouette}  \\
         & & {\small \it + Orig. BG} & {\small \it + White BG} \\
        \includegraphics[width=0.21\textwidth]{rws_images/blur_seg.png}
        & \includegraphics[width=0.21\textwidth]{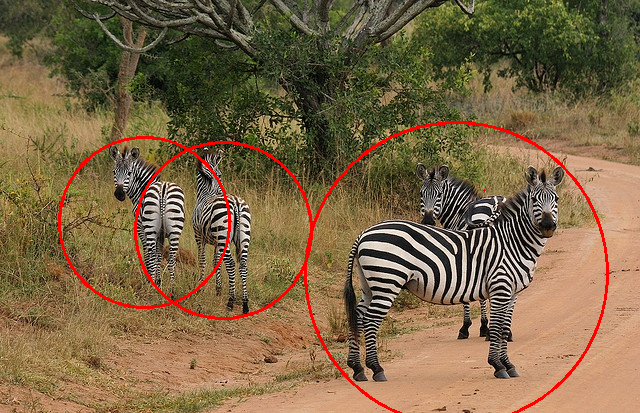}
        & \includegraphics[width=0.21\textwidth]{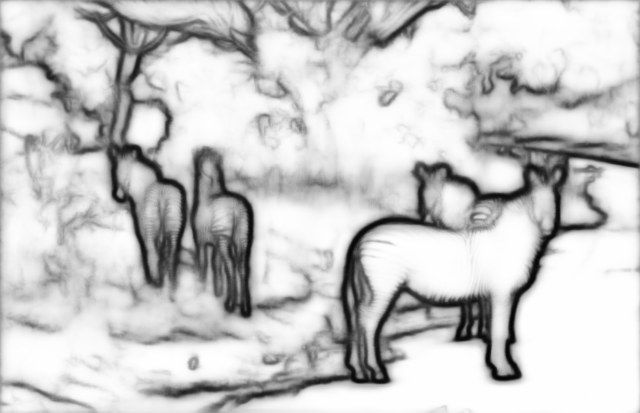}&
        \includegraphics[width=0.21\textwidth]{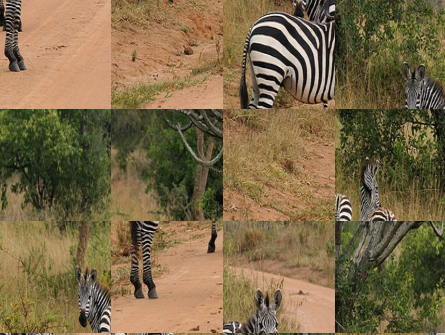}\\
                {\small \it Blur Reverse} & {\small \it Red Circle} & {\small \it Edge Map} & {\small \it Patch Shuffle}  \\  
    \end{tabular}
\end{center}
\caption{{\bf Background Transformations.} Samples illustrating image transformations that we consider in our analysis of performance in {\em visual}, {\em text} and {\em response} spaces of VLMs.\vspace{-0.2in} }
\vspace{-0.15in}
\label{fig:bg}
\end{figure}

\vspace{0.1in}
\noindent\textbf{Role of background context in VLMs.} In order to understand the role of background, we consider two cases where - 1) background is removed 2) object is masked. We notice that clean (black \& white) backgrounds (row 2 \& 3 in Table~\ref{tab:background}) improve performance considerably in the visual and VL proj spaces ($\sim$11-12\%) and relatively less in the response space ($\sim$4-5\%), pointing to loss of information as discussed earlier. This is likely because removing the background removes any distractors ({\em e.g.}, objects in the background), helping the VLMs focus on the object of interest. 
In the absence of a clear foreground (object is masked with a silhouette; rows 4 \& 5), 
we notice having background context (row 4) plays an important role and improves performance for all three spaces compared to row 5 (no background).
Overall, this points to the fact that removing background helps when object details are clear but 
when objects are masked, VLMs can find hints in the background to help with object recognition . This hypothesis is also confirmed by blurring which we discuss next. 

\vspace{0.1in}
\noindent\textbf{Visual prompting.} We now consider a reverse blur transformation where the background is blurred w.r.t the object \cite{fineprompt}. Blurring the background seems to be ``the best of both worlds'', helping VLMs focus on the object while still preserving some background context---leading to the best performance overall (row 6).  
Furthermore, even though the performance improvement is similar to the black \& white backgrounds for the visual and text spaces ($\sim$11-12\%), the increase in performance from blurring is higher for the response space ($\sim$7-8\%), pointing to the fact that some visual prompting techniques make up for the loss of information in response space and enhance performance more. Finally, we also experimented with adding a red circle to the object of interest \cite{red_circle}, with (row 7) and without background (row 8). The gains from the red circle in both cases are relatively smaller for all spaces.

\noindent\textbf{Role of Shape.} To investigate whether VLMs rely on the shape of object for recognition, we apply the edge maps and patch shuffle image transformations following\cite{shape_texture_v2,shape_texture_v1}. Edge maps (row 10) retains the object shape while removing other details such as texture, color and background, while patch shuffle (row 11) distorts the shape of the object. We observe a drastic drop in performance in the case of edge maps compared to a smaller drop for patch shuffle, suggesting that shape plays a relatively less important role in VLMs' object recognition. Finally, we note that for patch shuffle, the performance drop for the visual and text spaces is large while the response space is relatively more stable to such changes in high-performing VLMs (LLaVA, LLAVA-NEXT).

\subsection{Summary of Insights on Image Transformations}
\begin{enumerate}[leftmargin=*,topsep=4pt]
\setlength\itemsep{0em}

\item[\scriptsize \circled{1}] VLM's response space is most robust to corruptions, but posit this is largely due to deficiencies of information flow  (see \S\ref{sec:results}). VL projection space is least robust compared to other spaces, pointing to potential vulnerabilities. 

\item[\scriptsize \circled{2}] VLMs rely on both object information and background context to recognize objects and perform best when both are well-preserved. 

\item[\scriptsize \circled{4}] Visual prompting techniques improve performance dramatically for visual and VL projection spaces but the improvements are comparatively diminished for response space; showing another limitation of miss-alignment of information. 

\item[\scriptsize \circled{5}] We observe that VLMs, contrary to humans, rely on texture over shape information for their decision making. This is an important deficiency that may impact their generalization abilities. 
Inducing reliance on shape is of importance.


\end{enumerate}

\section{Conclusion}
\label{sec:conclusion}
In this paper, we set out to understand the limitations of prominent VLMs on fundamental visual tasks. We go significantly beyond the current benchmarks, which simply measure final performance, and construct a series of tests that probe which specific components of VLM design may be lacking. Our analysis design reveals that for most visual tasks (except spatial understanding), the knowledge can be seen as being preserved in the visual and intermediate VL projection space, however, it does not translate to the final VLM response space (especially for tasks like fine-grained recognition,
object counting, visual prompting). Overall, through our insights, we hope to encourage targeted efforts on reducing info. loss and translating knowledge to an accurate final VLM response.  



\vspace{0.05in}
\noindent
{\bf Acknowledgments.}
This work was funded, in part, by the Vector Institute for AI, Canada CIFAR AI Chairs, NSERC Canada Research Chair (CRC), and NSERC Discovery Grants. Resources used in preparing this research were provided, in part, by the Province of Ontario, the Government of Canada through CIFAR, the Digital Research Alliance of Canada\footnote{\url{ https://vectorinstitute.ai/\#partners}}, companies sponsoring the Vector Institute, and Advanced Research Computing at the University of British Columbia. Additional hardware support was provided by John R. Evans Leaders Fund CFI grant and Compute Canada under the Resource Allocation Competition award.

\section*{Limitations and Ethical Considerations}
\label{sec:limitations}
We study capabilities of well-known VLMs to better understand their design on basic visual tasks such as visual recognition, object counting and spatial skills. Extensions to more advanced 
capabilities such as segmentation and visual reasoning would be interesting directions of future work.  In any such analysis of existing models, one limitation is that it is challenging to calibrate for training data and augmentation techniques used in the various VLM models. We have put in our best efforts to objectively analyze all aspects of the experiments that could be under our control, however.

\vspace{0.05in}
\noindent
{\bf Societal Impact.} VLMs themselves can, and already do, have a significant societal impact. Biases and inaccuracies in these models can have profound  impacts on a broad spectrum of application domains. Our approach, which attempts to understand capabilities of such models and their shortcomings, can serve to mitigate some of these concerns by attributing the different modules in existing VLMs to specific capabilities. This may enable the possibility of carefully editing models to remove such biases and inaccuracies. We believe that our efforts of understanding the capabilities of these models at a module granularity is a first step to resolve the inaccurate or inappropriate behavior of such models in different applications. 


\bibliography{acl_latex}

\clearpage
\appendix
\section{Appendix}
\label{sec:appendix}
In this supplementary section, we discuss the following details, which could not be included in the main paper owing to space constraints:

\begin{itemize}
    \item Performance of VLM response on scaling LLM Decoder for coarse and fine-grained recognition (in continuation to Figure \ref{fig:scaling} of main manuscript)
    \item Implementation details regarding task prompts and probe design (in continuation to Section \ref{sec:method} of main manuscript)
    \item Implementation details of different background transformations we consider (in continuation to Section \ref{sec:analysis:background} of main manuscript)
    \item Details of VLM model variants that were used for our analysis (in continuation to Section \ref{sec:method} of main manuscript)
    \item More details about the data and classes we use for our experiments (in continuation to Section \ref{sec:method} of main manuscript)

\end{itemize}

\begin{table*}[t]
\caption{{\bf Effect of Scaling LLM Decoder on three spaces for PASCAL (coarse-grained) and Stanford Dogs (fine-grained) recognition}}
\small
\centering
\scalebox{0.65}{\begin{tabular}{l c cc>{\columncolor{gray!20}}c>{\columncolor{gray!20}}c cc>{\columncolor{gray!20}}c>{\columncolor{gray!20}}c cc>{\columncolor{gray!20}}c>{\columncolor{gray!20}}c}
\toprule
\multirow{2}{*}{VLM Method} & & \multicolumn{4}{c}{PASCAL VOC} & \multicolumn{4}{c}{Dogs} & \multicolumn{4}{c}{\bf Average} \\ 
\cmidrule(lr){3-6} \cmidrule(lr){7-10} \cmidrule(lr){11-14} 
& & Visual & VL proj & Res.(probe) & Res.(text)
& Visual & VL proj & Res.(probe) & Res.(text) 
& Visual & VL proj & Res.(probe) & Res.(text) \\  \midrule
InstructBLIP \cite{instructblip} & 7B & 98.0 & 98.6 & 95.0 & 70.0 & 93.6 & 92.6 &{28.5} &{12.0} &95.8 &95.6 &61.8 &41.0\\ 
InstructBLIP \cite{instructblip} & 13B & 98.0 & 98.6 & 96.5 & 66.0 & 93.6 & 92.8 &{46.2} &{26.8} &95.8 &95.7 &71.4 &46.4\\ \hline
{LLaVA-1.5} \cite{llava} & 7B & 96.8 & 95.8 & 91.1 & 90.2 & 91.9 & 88.7 & 25.9 & 19.9 &94.4 &92.3 &58.5 &55.1\\ 
{LLaVA-1.5} \cite{llava} & 13B & 97.0 & 96.6 & 95.1 & 93.2 & 91.9 & 90.0 & 37.3 & 30.6 &94.5 &93.3 & 66.2 &61.9 \\ 
\hline
LLaVA-NEXT \cite{llavanext}& 7B & 97.3 & 96.6 & 94.1 & 94.1 & 90.4 & 87.2 &37.5 &31.0 &93.9 &91.9 &65.8 &62.6\\ 
LLaVA-NEXT \cite{llavanext}& 13B & 97.3 & 96.2 & 95.7 & 95.8 & 90.0 & 86.5 &37.0 &36.9 &93.7 &91.4 &66.4 &66.4\\ 
\bottomrule
\label{scaling}
\end{tabular}}
\vspace{-5pt}
\end{table*}
\subsection{Scaling LLM Decoder}
\label{scaling_appendix}
In this section, we elaborate on the results showing effect of scaling language decoder of competitive instruction-tuned VLMs ({\em i.e.}, InstructBLIP, LLaVA, LLaVA-NEXT) on visual, text and response space (in continuation to the Figure \ref{fig:scaling} bar plot of main manuscript).
Table \ref{scaling} shows the results for PASCAL VOC (coarse-grained) and Stanford Dogs (fine-grained) datasets as well as the average performance to help observe overall trends. On average, we notice that a bigger LM decoder (i.e 13B vs 7B model variant) does improve performance to some extent, including in the alignment space, for most cases but major trends discussed for coarse/fine object recognition remain same. Furthermore, we notice that the improvements on increasing decoder size are relatively larger for weaker instruction-tuned models like InstructBLIP but gradually saturate for models like LLaVA-1.5 and LLaVA-NEXT which have stronger instruction tuning and better overall in terms of performance.
    

\subsection{Task Prompts and Evaluation of Response}
\label{task_prompts}
In this section, we discuss the details of our task prompts and design of our trainable probes and inference mechanism we use to analyse the knowledge in three spaces of a VLM. As discussed in main manuscript Section \ref{sec:method}, we consider different categories of VLM models ({\em i.e.}, Contrastive Multi-Encoder models (like, CLIP), Encoder-Decoder Generative models (like, BLIP2) and instruction-tuned models (like, LLaVA, InstructBLIP) which demand the use of different strategies to best evaluate their capabilities on visual understanding tasks, which we discuss below.

\begin{table*}[t]
\caption{{\bf Visual Corruptions}- Results for individual corruptions for VLMs in continuation to Table \ref{tab:robust} from main manuscript.  \vspace{-0.15in}}
\label{tab:visual_corruption_v1}
\small
\centering
\resizebox{1.0\textwidth}{!}{
\begin{tabular}{l cc>{\columncolor{gray!20}}c>{\columncolor{gray!20}}c cc>{\columncolor{gray!20}}c>{\columncolor{gray!20}}c cc>{\columncolor{gray!20}}c>{\columncolor{gray!20}}c}
\toprule
\multirow{2}{*}{Transformation} &  \multicolumn{4}{c}{LLaVA-NEXT \cite{llavanext}} & \multicolumn{4}{c}{LLaVA-1.5 \cite{llava}} & \multicolumn{4}{c}{InstructBLIP \cite{instructblip}} \\ 
\cmidrule(lr){2-5} \cmidrule(lr){6-9} \cmidrule(lr){10-12} 
& Visual & VL proj & Resp. (probe) & Resp.(text)
& Visual & VL proj & Resp. (probe) & Resp.(text)
& Visual & VL proj & Resp. (probe) & Resp.(text)\\  \midrule 

Snow    & 
        94.0 & 
        93.3 & 
        92.8 & 
        92.9 & 
        95.6 & 
        94.5 & 
        92.0 & 
        91.6 & 
        96.7 & 
        97.1 & 
        92.7 & 
        70.5 \\ 
        
Uniform noise    & 
        90.0 & 
        88.0& 
        87.2 & 
        87.2& 
        90.5 & 
        88.2 & 
        87.5 & 
        87.0 & 
        94.7 & 
        95.5 & 
        88.5 & 
        69.5 \\ 
Gaussian Noise & 
        85.5 & 
        84.3 & 
        82.7& 
        83.0 & 
        88.0 & 
        86.0 & 
        85.0 & 
        85.0 & 
        94.3 & 
        95.2 & 
        89.3 & 
        68.0   \\
Shot Noise & 
        89.0 & 
        86.4 & 
        85.7 & 
        85.8 & 
        89.5 & 
        88.5 & 
        86.5 & 
        86.3 & 
        95.7 & 
        96.1 & 
        92.5 & 
        69.0 \\ \hline
Impulse Noise & 
        87.5 & 
        85.5 & 
        84.8 & 
        85.0 & 
        87.8 & 
        87.0 & 
        85.5 & 
        85.5 & 
        95.1 & 
        95.8 & 
        90.5 & 
        68.0 \\   
Motion blur  & 
        92.2 & 
        92.0 & 
        90.5 & 
        90.5 & 
        93.2& 
        91.9 & 
        90.0 & 
        90.0 & 
        95.6 & 
        95.8& 
        89.5 & 
        70.5 \\
Defocus blur & 
        89.5 & 
        88.0 & 
        86.7 & 
        86.8 & 
        91.0 & 
        89.2 & 
        87.0 & 
        86.4 & 
        95.2 & 
        95.8& 
        89.6 & 
        71.0 \\ \hline \hline
 Average Corrupt. & 
        89.7 & 
        88.2 & 
        87.2 & 
        87.3 & 
        90.8 & 
        89.3 & 
        87.7 & 
        87.4 & 
        95.3 & 
        95.9& 
        90.5 & 
        69.5 \\ \hline
Original    &97.3  &96.6 &94.1 &94.1 &96.8  &95.8  &91.0 &90.2  &98.0  &98.6 &95.0  &70.0 \\ \hline
\bottomrule
\end{tabular}}

\end{table*}

\begin{table}[t]
\caption{{\bf Visual Corruptions}- Results for individual corruptions for VLMs in continuation to Table \ref{tab:robust} from main manuscript.\vspace{-0.15in}}
\label{tab:visual_corruptions_v2}
\small
\centering
\resizebox{1.0\textwidth}{!}{
\begin{tabular}{l cc>{\columncolor{gray!20}}c>{\columncolor{gray!20}}c}
\toprule
\multirow{2}{*}{Transformation} &  \multicolumn{4}{c}{BLIP-2 \cite{li2023blip2}} \\ 
\cmidrule(lr){2-5}
& Visual & Text & Resp. (probe) & Resp. (text)\\  \midrule 

Snow    & 
        96.2 & 
        96.8 & 
        65 & 
        40.0 \\
Uniform noise    & 
        94.7 & 
        94.7 & 
        55.0 & 
        34.5 \\
Gaussian noise    & 
        93.9 & 
        94.4 & 
        54.0 & 
        38.6 \\
Shot noise    & 
        95.7 & 
        96.1 & 
        54.0 & 
        38.5 \\

Impulse noise    & 
        95.0 & 
        94.7 & 
        54.5 & 
        37.5 \\

Motion blur    & 
        95.8 & 
        96.0 & 
        55.0 & 
        38.4 \\

Defocus blur    & 
        95.2 & 
        95.4 & 
        65.5 & 
        39.9 \\\hline\hline
Average Corrupt.    & 
        95.2 & 
        95.4 & 
        57.5 &
        38.2 \\\hline
Original    &  98.0  & 99.0 & 57.6 & 38.0   \\ \hline

\bottomrule
\end{tabular}}

\end{table}
\begin{itemize}
\itemsep0em 
\item\textbf{CLIP, CoCa and ALBEF:} Given these are multimodal contrastive models with a separate encoder for visual and text modalities, we evaluate the response space performance using standard zero-shot inference procedure with the help of cosine similarity in the shared embedding space. Specifically, we use similarity visual features (from visual encoder) and text prompt features (from text encoder) for inference and choose the class prompt with the maximum similarity as the predicted answer. We use the default prompts used for CLIP such as {\small \tt [A photo of a \textbf{cat}, A photo of a \textbf{dog}…..]} for \textbf{object recognition} task, {\small \tt [There is \textbf{one} cat, There are \textbf{two} cats, There are \textbf{three} cats ..]} for \textbf{object counting} task and {\small \tt [Dog is to \textbf{left} of cat, Dog is to the \textbf{right} of cat, Dog is \textbf{below} cat …]}  for \textbf{spatial orientation} understanding tasks.

\item \textbf{LLaVA-1.5, LLaVA-NEXT, BLIP2 and InstructBLIP} For VLMs which generate textual response, we used the standard procedure as followed in previous work \cite{finer, distribution}, where a given visual task, we prompt the model with a question and choices for the task, and ask it to select (generate) the right answer from the choices. For the task prompts, we follow the prompts similar to those used in previous work by \cite{distribution}. For example, to evaluate object recognition for LLaVA model, we use the prompt {\small \tt What is the central object in the image out of the following list of choices? Make sure to answer in one word. Choices: [cat, dog, frog ...]}. Note that  LLaVA-1.5 and LLaVA-NEXT are specifically instruction-tuned to understand if the user wants it to answer in one word which is why we incorporate the phrase ``Make sure to answer in one word". For BLIP2 and InstructBLIP model we use the same prompt {\em i.e.}, {\small \tt What is the central object in the image out of the following list of choices? Choices: cat, dog, frog ...} without the phrase ``Make sure to answer in one word" since these models do not understand answering in one word and have continuous outputs. We found that adding the phrase reduced performance by distracting these VLMs model, thus we remove it as we want to fairly and best evaluate the response space capabilities of all VLMs. 
We follow a similar strategy for counting and spatial orientation tasks where we ask {\small \tt How many dogs are there in the image? Choices: 1,2,3,4} and {\small \tt What is the spatial location of dog relative to cat? Choices: left, right, above, below} and expect it to answer with the total number of objects or the spatial orientation between objects.

\textbf{Evaluation of textual VLM response} - As detailed in the main manuscript, to facilitate fair comparison with visual and text feature probes and comprehensive evaluation of response space, we use 2 strategies - 1) training linear probe atop of output response tokens from the language decoder (we refer to this LM Dec.) and 2) evaluating the textual response of the VLM for a given task using matching between the textual VLM response and the ground truth class label (we refer to this Resp.). We discuss the latter Resp. strategy here and discuss the former LM Dec. strategy in next Section \ref{probe_design} (probe design).Specifically, in order to evaluate whether the VLM answered with the correct class from the choices, we use matching between the textual VLM response and
the ground truth class label  similar to previous work on evaluating VLMs \cite{finer}, where we check which classname in the choices matches the best with the VLM output with the help of regex based matching and fuzzy matching to make sure we match the output to the correct classname irrespective of punctuation, capitalization or style of output (one word vs long/continuous output). 

\end{itemize}

\subsection{Control Task for Probes} 
\label{probe_design}
\noindent
When probing is used as a mechanism to test specific properties encoded by representations, it is essential to make sure that success on the probing task implies the representation encodes the properties useful for the task rather than they were learned by the probe itself. We follow previous work and report the performance on a 
control task in which the gold labels have been randomly shuffled  \cite{zhang-bowman-2018-language, control_task}.  We expect a good probe to have low accuracy on the control task. Table~\ref{table:control} shows that the performance of all three spaces of the state-of-the-art LLaVA-NEXT model on the control task for all tasks (coarse- and fine- grained classification, counting, and spatial arrangement) is poor;   supporting the validity of our probing mechanism.

\begin{table}[t]
    \scriptsize
    \setlength{\tabcolsep}{4pt}
    \centering
    \begin{tabular}{@{}lccc>{\columncolor{gray!20}}c>{\columncolor{gray!20}}c@{}}
    \toprule
    \multirow{2}{*}{VLM Method} & & \multicolumn{3}{c}{Control Task Performance}  \\ 
    \cmidrule(lr){3-5}
    & & Visual & Text & Resp. (Probe) \\  \midrule 
    Classification (coarse-grained)  & & 10.4 & 10.6 & 15.5  \\ 
    Classification (fine-grained) & & 4.7 & 6.1 & 4.9  \\ 
    Spatial & &  25.4 & 23.6 & 22.5 \\
    Counting & & 23.8 & 25.7 & 9.0 \\
    
    \bottomrule
\end{tabular}
\caption{\textbf{Control Task for Probes}: We show performance for control task for Coarse-grained classification (PASCAL), Fine-Grained (Stanford Dogs), Counting and Spatial Tasks (PaintSkills). The performance is low for all three spaces supporting the validity of our probing mechanism} 
\label{table:control}
\end{table}


Given that we treat each visual task as classification and the datasets we use for visual tasks have ground-truth labels for each image ({\em i.e.}, object class labels, spatial orientation labels and object counting labels), we use simple \textit{accuracy} (on validation set) as the metric for evaluation on object recognition, spatial understanding, and object counting tasks; following  previous work \cite{DallEval}. Given our linear probes are lightweight we use 1 A100 GPU for all our experiments.

\subsection{Visual Corruption Results}
\label{visual_corr_appendix}
In the main manuscript, we reported only the average performance across all corruptions (i.e Corrupt.) Table \ref{tab:robust} and compared it to the original performance (without corruptions) i.e Orig, owing to space constraints. Here, we provide the tables of performances for each corruptions (Table \ref{tab:visual_corruption_v1} and Table \ref{tab:visual_corruptions_v2})
\subsection{Implementation: Background Transforms}
\label{background_implementation}
In this section, we discuss the implementation of background transforms used in Sec.\ref{sec:analysis:background}.  Specifically, we refer to recent work on background transformation \cite{background} which provides COCO images and corresponding masks for foreground objects extracted using FastSAM. The presence of foreground object masks for each image helps us separate the object foreground and background and apply simple transformation techniques to get the {\small \tt white background}, black {\small \tt background changes} and {\small \tt silhouettes}. For {\small \tt reverse blur}, we follow \cite{fineprompt} and blur the background with the help of gaussian filter. To generate {\small \tt edge images} which retain global object shapes and remove texture related information, we use the technique previously used by \cite{shape_texture_v2} to get edge maps with the help of RCF \cite{liu2019richer} based edge detection.  For the {\small \tt red circle }transformation, we simply us contour detection for the object of interest and draw the minimum enclosing circle using OpenCV library. Finally, for {\small \tt patch shuffle}, we use pytorch transformations to divide image into 4x4 grid and shuffle the patches.
\subsection{Experimental Setup Details}
\label{sec:setup}
{\bf Choice of VLM variants.} Here, we elaborate on the details of specific model variants we use for our analysis.


\begin{itemize}
\itemsep0em 
\item[-] {\bf \em CLIP} \cite{clip}. Contrastive Language-Image Pre-Training (CLIP) consists of separate image and language encoders trained jointly using a contrastive objective that aligns the two latent representations. From the perspective of analysis in this paper, CLIP has the {\em visual latent} which can be linear probed and the {\em response space} performance can be obtained by image-text similarity based matching in the latent space; it does not have an projection/alignment ({\em text space}). 

For the experiments in this paper we use a variant with {\tt ViT-L-14-336} backbone as visual encoder. Our choice is motivated by the fact that this variant is similar to the encoders of other VLMs we consider for our analysis (in terms of architecture and parameter size) and hence allows for fair comparison.
\item[-] {\bf \em CoCa} \cite{coca}.-
Contrastive Captioners are Image-Text Foundation Models (CoCa) uses  contrastive loss (similar to CLIP) jointly in combination with a captioning (generative) loss making it a hybrid contrastive $+$ encoder-decoder generative model. It has {\em visual latent} and the {\em response space} performance can be obtained by image-text similarity based matching in the latent space similar to CLIP. 

For the experiments in this paper we use a variant with {\tt ViT-L-14} backbone as visual encoder sinces it is similar to the encoders of other VLMs we consider for our analysis (in terms of architecture and parameter size) and hence allows for fair comparison.
\item[-] {\bf \em ALBEF} \cite{ALBEF}.- ALign BEfore Fuse first encodes the image and text independently and adds an additional multimodal encoder to fuse the image features with the text features through cross- modal attention. It uses a combination of image-text contrastive (ITC), image-text matching (ITM) and masked language modeling (MLM) losses.

For the experiments in this paper we, we use the standard LAVIS library implementation and evaluate the {\tt albef-feature-extractor} {\tt (base)} model variant. We use the same inference procedure as used for CLIP and CoCa models where we probe the visual latent features and {\em response space} performance can be obtained by image-text similarity based matching.

\item[-] {\bf \em BLIP 2} \cite{li2023blip2} and {\bf \em InstructBLIP \cite{instructblip}}. 
Bootstrapping Language-Image Pre-training (BLIP) is a popular VLM model that leverages pre-trained image encoder (CLIP) and a LLM decoder ({\em e.g.}, OPT, FlanT5) and learn to align the image representation to the expected input of the LLM decoder. The alignment mechanism takes the form of a small transformer-based neural network (Q-former) that is trained with a variety of losses. While standrad BLIP2 model keeps the image encoder and language decoder frozen, the instruction fine-tuned InstructBLIP model variant uses high quality image-language pairs to instruction fine-tune the performance of the full architecture. 

For our analysis  we use variants of BLIP2 and InstructBLIP that are closest in size to other VLM models we consider. For BLIP, this means {\tt VIT-G-14} as the visual encoder and {\tt OPT} as the language decoder. For InstructBLIP, we use the model with {\tt VIT-G-14} as the visual encoder and {\tt Vicuna} as the language decoder.

\item[-] {\bf \em LLaVA and LLaVA-NEXT} \cite{llava}.
Large Language and Vision Assistant (LLaVA) is a popular and one of the most competitive VLM model for general-purpose visual and language understanding. It uses language-only GPT-4 to generate multimodal language-image instruction-following data and implements carefully designed training procedures and architectural choices that results in impressive multimodal chat abilities. 
In our analysis, we use the {\tt LLaVA-1.5} and {\tt LLaVA-NEXT} model variants which shows excellent instruction following and multimodal chat capabilities across different VLM benchmarks. The variant we consider uses {\tt CLIP ViT-L-14} backbone as visual encoder (same as the CLIP model above) and {\tt Vicuna} as LLM decoder. LLaVA-NeXT has improved reasoning, OCR, and
world knowledge compared to LLaVA-1.5, thanks to better instruction tuning on high quality user instruct data. It achieves the
best performance among open-source LMMs; exceeds Gemini Pro
and outperforms Qwen-VL [1] on several benchmarks \cite{llavanext}.

\end{itemize}

\subsection{Dataset Statistics and Class Details}
\label{dataset_appendix}
In this section, we elaborate on the details of object classes and dataset statistics we use for the different tasks we consider in our paper. Note that as discussed before, we train our probes on the training data and test it on the val/test split following standard protocol. 

Specifically, for coarse-grained recognition we consider the the PASCAL VOC with object classes {\small \tt ['aeroplane', 'bicycle', 'bird', 'boat', 'bottle', 'bus', 'car', 'cat', 'chair', 'cow', 'diningtable', 'dog', 'horse', 'motorbike', 'person']}and,
\\PaintSkills dataset with object classes {\small \tt ['person', 'airplane', 'bicycle', 'dog', 'boat', 'car', 'fire hydrant', 'stop sign', 'umbrella', 'bench', 'suitcase', 'traffic light', 'bird', 'bear', 'potted plant']}.\\ For fine-grained recognition we consider the CUB data with classes
{\small \tt ['black footed albatross', 'laysan albatross', 'sooty albatross', 'groove billed ani', 'crested auklet', 'least auklet', 'parakeet auklet', 'rhinoceros auklet', 'brewer blackbird', 'red winged blackbird', 'rusty blackbird', 'yellow headed blackbird', 'bobolink', 'indigo bunting', 'lazuli bunting']} and\\ Stanford Dogs dataset with object classes{\small \tt ['chihuahua', 'japanese spaniel', 'maltese dog', 'pekinese', 'shih-tzu', 'blenheim spaniel', 'papillon', 'toy terrier', 'rhodesian ridgeback', 'afghan hound', 'basset', 'beagle', 'bloodhound', 'bluetick', 'black-and-tan coonhound']}

As mentioned previously, for counting and spatial recognition task we use object counts {\small \tt[1,2,3,4]} and spatial relations {\small \tt[left,right,above,below]}

We follow standard splits of PaintSkills dataset \cite{DallEval} with  23,250/21,600/13,500 and 2,325/2,160/2,700 scenes
for train and test splits of object recognition/object counting/spatial relation understanding skills, respectively. For Pascal VOC/Stanford Dogs/CUB datasets we have 2100/1500/500 and 2100/1300/400 for train and test splits, respectively, with the aforementioned object classes.

\end{document}